\documentclass[runningheads]{llncs}
\usepackage[T1]{fontenc}
\usepackage{indentfirst}
\usepackage{graphicx}
\usepackage{booktabs}
\usepackage{graphicx}
\usepackage{tabularray}
\usepackage{enumitem}
\usepackage{hyperref}
\usepackage[capitalize]{cleveref}
\usepackage{xcolor}
\usepackage{listings}
\usepackage{xparse}
\lstdefinestyle{mystyle}{
    language=SQL,
    keywordstyle=\color{blue}\bfseries\ttfamily,
    basicstyle=\ttfamily,
    breakatwhitespace=false,         
    breaklines=true,                 
    captionpos=b,                    
    keepspaces=true,                 
    showspaces=false,                
    showstringspaces=false,
    showtabs=false,                  
    tabsize=2
}
\crefname{figure}{Fig.}{Figs.}
\crefname{table}{Table}{Tables}
\crefname{equation}{Eq}{Eqs}

\newcommand\blfootnote[1]{%
  \begingroup
  \renewcommand\thefootnote{}\footnote{#1}%
  \addtocounter{footnote}{-1}%
  \endgroup
}

\begin{document}
%
% \title{Prompt Tuning for Natural Language to SQL with Embedding Fine-Tuning and Retrieval Augmented Generation}
\title{Prompt Tuning for Natural Language to SQL with Embedding Fine-Tuning and RAG}
% \thanks{}
\titlerunning{Prompt Tuning for NL2SQL with Embedding Fine-Tuning and RAG}
\author{Jisoo Jang \and 
Tien-Cuong Bui \and Yunjun Choi \and Wen-Syan Li
}
\authorrunning{Jisoo et al.}
\institute{Graduate School of Data Science, Seoul National University, Seoul, South Korea\\ 
\email{\{simonjisu,cuongbt91,datajun77,wensyanli\}@snu.ac.kr}}

\maketitle              % typeset the header of the contribution
\blfootnote{Presented at the Workshop on Robust ML in Open Environments (PAKDD 2024)}
\begin{abstract}

This paper introduces an Error Correction through Prompt Tuning for NL-to-SQL, leveraging the latest advancements in generative pre-training-based LLMs and RAG. Our work addresses the crucial need for efficient and accurate translation of natural language queries into SQL expressions in various settings with the growing use of natural language interfaces. We explore the evolution of NLIDBs from early rule-based systems to advanced neural network-driven approaches. Drawing inspiration from the medical diagnostic process, we propose a novel framework integrating an error correction mechanism that diagnoses error types, identifies their causes, provides fixing instructions, and applies these corrections to SQL queries. This approach is further enriched by embedding fine-tuning and RAG, which harnesses external knowledge bases for improved accuracy and transparency. Through comprehensive experiments, we demonstrate that our framework achieves a significant 12 percent accuracy improvement over existing baselines, highlighting its potential to revolutionize data access and handling in contemporary data-driven environments.

\keywords{NL-to-SQL \and Embedding Fine-Tuning \and Large Language Models \and Retrieval-Augmented Generation}
\end{abstract}
\section{Introduction}\label{c:introduction}

Natural language interfaces (NLIs) offer a convenient means for querying and interacting with various data types. Increasingly, individuals are using everyday language to access online and offline information. For instance, people might inquire about today's weather or discuss tomorrow's agenda on their personal calendars with NLIs. A notable example is Moveworks, a company that primarily develops an AI copilot to boost employee productivity. This technology enables employees to effortlessly access personal or company data using text, which employs natural language. As generative AI continues evolving, natural language interfaces seamlessly integrate into our everyday lives.

Natural Language Interfaces to Databases (NLIDBs) \cite{androutsopoulos1995natural} facilitate data access in the Natural Language Processing (NLP) domain by allowing users to query databases using natural language. These systems aim to translate a natural language question $q_{nl}$ into an equivalent SQL expression $q_{sql}$ (NL-to-SQL) in a given database system $D$. Various approaches have been explored for NLIDBs. Early methods like LUNAR \cite{woods1972lunar} and Athena \cite{saha2016athena} employed rule-based systems to parse or map NL queries to an ontology syntactically and then to SQL. With advancements in NLP and neural networks, intermediate representation systems transform NL queries into vector representations before generating SQL. For instance, SQLova \cite{hwang2019comprehensive} and HydraNet \cite{lyu2020hybrid} predict SQL keyword sections, while RATSQL \cite{wang2019rat} and BRIDGE \cite{lin2020bridging} utilize encoder-decoder frameworks for SQL generation.

Recent advancements in generative pre-training-based large language models (LLMs) \cite{radford2019language}\cite{brown2020language} address NL-to-SQL challenges. These models, trained on vast internet-sourced data, excel in understanding human language. They operate through sequence-to-sequence tasks, where users provide a prompt to generate conditional answers. Research shows that LLMs' effectiveness in NL-to-SQL tasks is significantly influenced by the design of these prompts. Notably, including schema and foreign critical information in prompts led to a 51.6\% improvement in execution accuracy on the Spider development dataset \cite{rajkumar2022evaluating}\cite{yu2018spider}, a benchmark for large-scale NL-to-SQL tasks. 

Two fundamental approaches for improving LLM-based NL-to-SQL are fine-tuning \cite{li2022resdsql}\cite{scholak2021picard}, adapting LLMs to structured tabular data, and inference-only \cite{liu2023pre} with either prompt engineering or few-shot in-context learning. For instance, Pourreza et al. \cite{pourreza2023din} improved the generation performance by decomposing problems and employing chain-of-thought \cite{wei2022chain} with few-shot prompting. Despite numerous advancements, the few-shot approach depends on the quality and quantity of examples \cite{song2205comprehensive}. The Retrieval-Augmented Generation (RAG) approach \cite{lewis2021retrievalaugmented} can mitigate the few-shot approach drawbacks, which enhances LLMs by integrating facts from external knowledge bases. It ensures access to the most current and accurate information and offers users transparency in the model's generative process. However, optimizing content augmentation within LLMs' constrained input size remains a complex issue in the NL-to-SQL field.

\begin{figure*}[ht]  
\vspace{-12pt}
\centerline{\includegraphics[width=\textwidth]{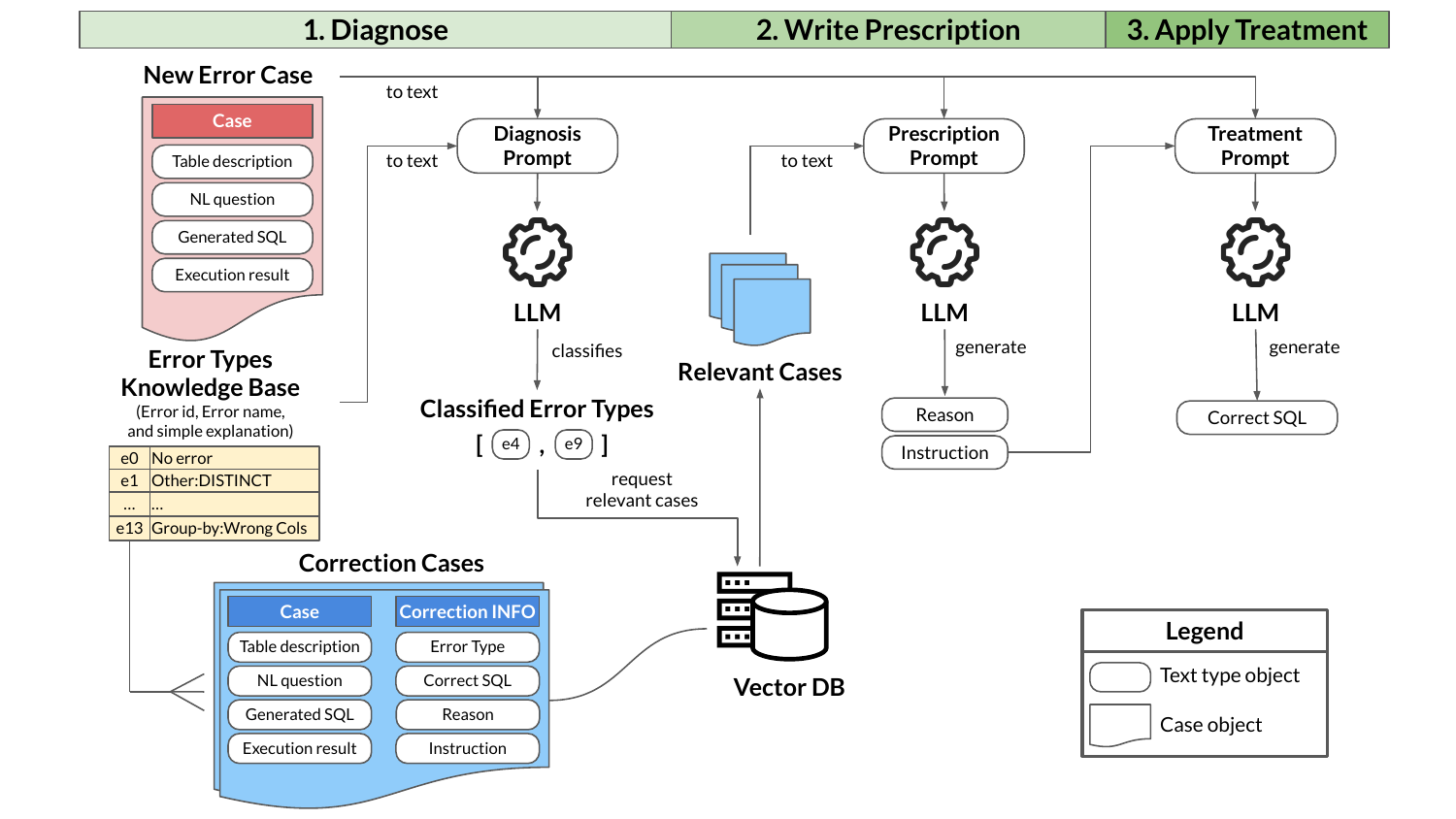}}
\vspace{-5pt}
\caption{Overview of Error Correction through Prompt Tuning (ECPT)}
\label{fig:framework}
\vspace{-8pt}
\end{figure*}

This paper proposes an \textbf{E}rror \textbf{C}orrection framework based on \textbf{P}rompt \textbf{T}uning with embedding fine-tuning and RAG for NL-to-SQL. The motivation for this work stems from medical diagnosis processes. As depicted in Fig.~ \ref{fig:framework}, our framework incorporates an error correction process that diagnoses error types, finds the reasons, provides fixing instructions based on RAG, and applies the instructions to fix SQL errors. To improve retrieval processes, we fine-tune a pre-trained Sentence Transformer model with a customized error correction dataset.
We evaluate the efficiency and correctness of our framework through extensive experiments. Experimental results demonstrate that our framework achieves a 12\% accuracy improvement compared to baselines.

% % The following contents...
The rest of the paper is organized as follows: \cref{c:background} introduces background. \cref{c:erroranalysis} presents error analysis. \cref{c:method} explains our framework. \cref{c:experiments} shows our experiment results. Finally, \cref{c:conclusion} summarizes the findings and discusses the limitations and future works.

\section{Background}\label{c:background}

\subsection{Large language model (LLM)}
LLMs like GPT-3.5 and GPT-4, PaLM \cite{chowdhery2022palm}, and LLaMa \cite{touvron2023llama} achieve comprehensive language understanding and generation via pre-training with an enormous amount of data and fine-tuning with RLHF \cite{ouyang2022training}. Additionally, domain-specific LLMs like BloombergGPT \cite{wu2023bloomberggpt}, trained with domain-specific and proprietary data, are also prevalent. LLMs excel in numerous NLP tasks since they can understand diverse input contexts through billions of trained parameters.

\subsection{Fine-tuning Language Models}

Foundation models can be fine-tuned for specific tasks through primary methods: \textbf{model fine-tuning},  \textbf{prompt tuning}, and \textbf{prompt engineering}. Model fine-tuning involves adjusting the pre-trained model's weights using supplementary labeled data, which can be both time-intensive and costly. Conversely, prompt tuning or prompt engineering \cite{liu2023pre} is more cost-efficient. In prompt tuning, AI-generated numerical sequences, known as \textbf{soft prompts}, alter a separate tuneable model's embedding space to guide LLM to perform specific tasks. Additionally, human engineers can employ \textbf{hard prompts} or engineered prompts, which are instructions or examples integrated with the input prompt, to steer the model toward specific tasks. These hard prompts, considered few-shot learning examples, are more interpretable but generally less effective than the AI-generated soft prompts. While soft prompts offer better performance, they lack interpretability, resembling a `black box' similar to deep learning models, whereas hard prompts, being engineer-generated, are easier to understand.

\subsection{Retrieval-Augmented Generation}

Fine-tuning a model alone often fails to equip it with the comprehensive knowledge necessary to answer particular, contextually evolving questions. Since LLMs are trained with Internet data, they may suffer from inconsistencies and outdated information. \textbf{Retrieval-Augmented Generation (RAG)} \cite{lewis2021retrievalaugmented}, enhancing LLMs by retrieving facts from external knowledge bases is a solution for these issues. It reduces the need for continuous retraining and fine-tuning of new data, resulting in better costs of maintaining high-quality Q\&A performance. Additionally, RAG can be used together with fine-tuning approaches to maximize the benefits of both approaches.

% In this paper, we use soft prompts (changing the embedding vector space), hard prompts (i.e., error type cases), and RAG to achieve a higher NL-to-SQL error correction rate.
\section{Error Analysis}\label{c:erroranalysis}

To gain a clearer insight into the limitations of LLMs in a zero-shot scenario, we randomly selected a subset of 16 databases and 959 NL questions in the Spider \cite{yu2018spider} training dataset. The basic prompt was utilized in conjunction with foreign keys to assist LLMs in formulating accurate responses to NL questions, as suggested in \cite{gao2023text}. After executing the generated queries, we classified the execution results as follows: 
\begin{itemize}[label=$\bullet$]
\item \textbf{Success:} The SQL executed successfully compared to the ground-truth queries.
\item \textbf{Execution Error:} An SQL system error occurs when executing the generated SQL. 
\item \textbf{Empty Table:} The SQL result returns an empty result. Usually, it happens when there are no matching values in the WHERE clause. 
\item \textbf{Undesired Result:} The SQL executed well but did not match the user's expectation compared to the ground-truth queries. 
\end{itemize}

By double-checking the execution results and the generated queries manually, we excluded 14 questions that the natural language question did not match the ground-truth query. For example, it is hard to consider selecting \lstinline|rID| from the table in the question `Find the title and star rating of the movie that got the least rating star for each reviewer.' The SELECT clause in the ground-truth query was \lstinline|SELECT T2.title, T1.rID, T1.stars, min(T1.stars)|. 

\begin{figure*}[h]  
\centerline{\includegraphics[width=0.6\textwidth]{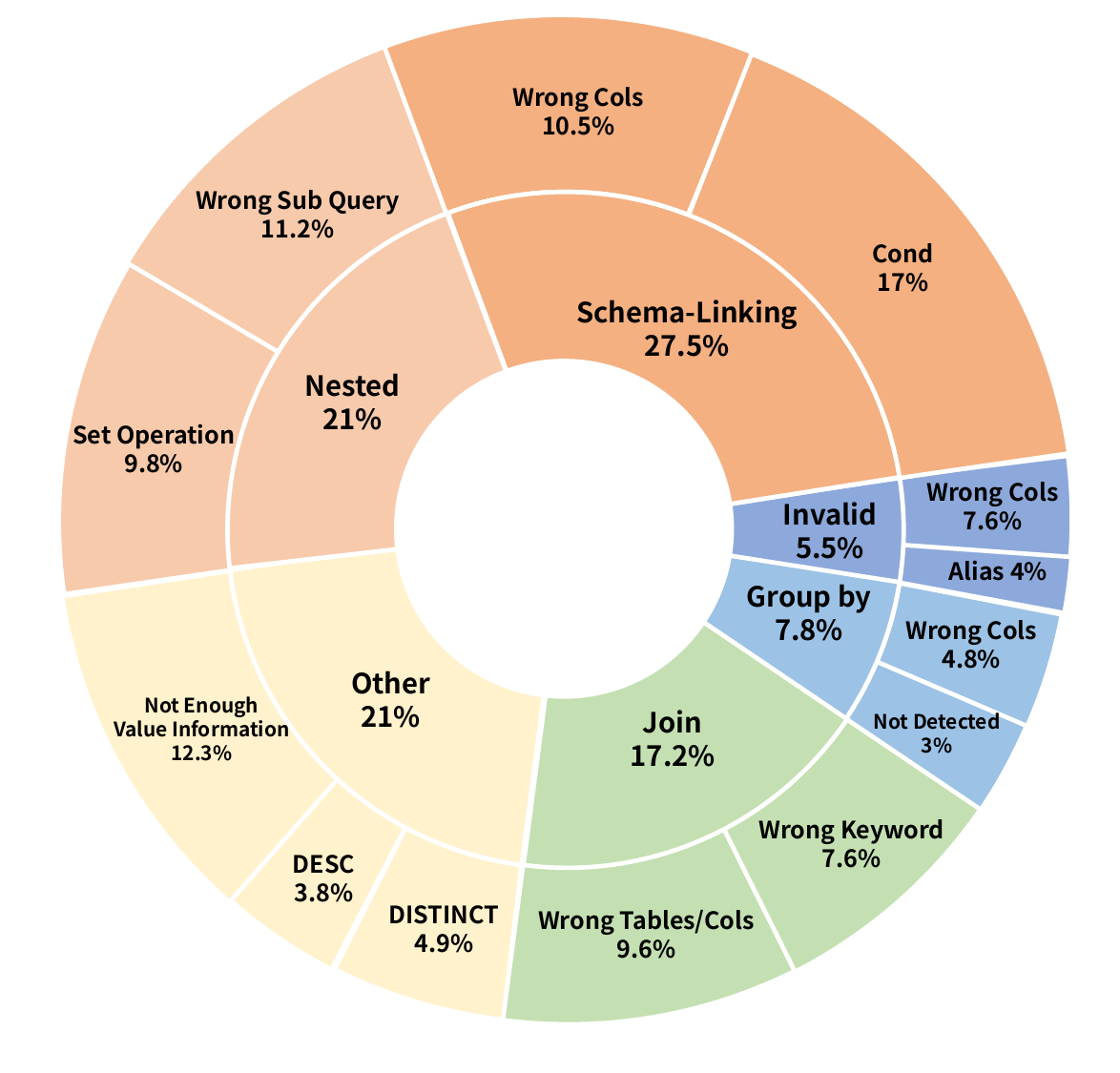}}
\vspace{-10pt}
\caption{Statistics of error classification}
\label{fig:error}
\vspace{-10pt}
\end{figure*}

To further study the execution results, our team manually classified them into six categories, similar to \cite{pourreza2023din} but with some modification, as shown in Fig.~\ref{fig:error}. One of the different categories is `Other:Not Enough Value Information'. It has been discovered that many SQL queries lack value information, resulting in an empty table. E.g., in the question `What are all the different food allergies?' and the ground-truth query of WHERE clause is \lstinline|WHERE allergytype = "food"|. LLM generates \lstinline|"Food"| instead of \lstinline|"food"|. 

Since LLMs are trained on extensive data and code for general purposes but not trained on tasks on structured tabular data, they might struggle with knowledge of enterprise-scaled data lakes. They can craft queries from basic table descriptions containing only the table, column names, and foreign keys but often miss crucial value information, leading to inaccuracies. This guides us in developing an idea to construct a knowledge base for error correction and utilizing RAG to solve the problem. 

When categorizing, we note reasons for LLM-generated failures and suggest instructions. Ambiguous SQL is labeled as a failure against the ground-truth standard. For example, the question `Which major has the most number of students?' yields two similar but different SQL queries: ground-truth (\lstinline|SELECT major|) and generated query (\lstinline|SELECT Major, COUNT(*) AS StudentCount|).  Although both are technically correct, we set the ground-truth query as the desired answer to the NL question to resolve these ambiguities.

All the categories are described in Table \ref{tab:errtype}. Detailed descriptions of each category are in the supplementary material.

\section{Methodology}\label{c:method}

\textbf{Error Correction through Prompt Tuning (ECPT)} is a three-step approach for fixing SQL query errors. It starts with the LLM identifying error types using a Diagnosis Prompt. Relevant cases are then fetched from a vector database. The LLM explains the error and generates instructions on correcting it via the Prescription Prompt. Finally, the Treatment Prompt generates the corrected SQL with instructions. 
It is similar to how doctors make a diagnosis and try to treat a patient. Given error information, we decompose the error correction process into three steps as shown in Fig.~\ref{fig:framework}: (1) Diagnose, (2) Write Prescription, and (3) Apply Treatment.

\subsection{Terminology Definitions}

The term \textbf{case} refers to the outcome of zero-shot SQL generation and execution, encompassing a table description (including table, column names, and foreign keys), a natural language question, generated SQL, and its execution result. This information, formatted as structured text, aids in prompt creation or vector conversion for similarity searches in a vector database. 

\begin{table}[!hb]
\centering
\caption{Error Types}
\label{tab:errtype}
\resizebox{\linewidth}{!}{%
\begin{tblr}{
  row{1} = {c},
  cell{2}{1} = {c},
  cell{3}{1} = {c},
  cell{4}{1} = {c},
  cell{5}{1} = {c},
  cell{6}{1} = {c},
  cell{7}{1} = {c},
  cell{8}{1} = {c},
  cell{9}{1} = {c},
  cell{10}{1} = {c},
  cell{11}{1} = {c},
  cell{12}{1} = {c},
  cell{13}{1} = {c},
  cell{14}{1} = {c},
  hline{1,15} = {-}{0.08em},
  hline{2} = {-}{0.05em},
}
\textbf{Error ID} & \textbf{Error Name} & \textbf{Short Explanations} \\
e1 & Other:DISTINCT & Didn’t use or use keyword DISTINCT properly. \\
e2 & Other:DESC & Didn’t use or use keyword DESC properly. \\
e3 & Other:Not Enough Value Information & Wrong value in the WHERE clause. \\
e4 & Schema-Linking:Wrong Cols & Unnecessary or wrong columns in SELECT clause refer to question. \\
e5 & Schema-Linking:Cond & Missing or used wrong logic in the conditions. \\
e6 & Nested:Wrong Sub Query & Unnecessary or wrong sub query. \\
e7 & Nested:Set Operation & Didn’t used set operation. \\
e8 & Join:Wrong Tables/Cols & Joined unnecessary or wrong tables or columns. \\
e9 & Join:Wrong Keyword & Didn’t use JOIN keyword where it should be used or misuse LEFT/RIGHT JOIN. \\
e10 & Invalid:Wrong Cols & Use columns that do not exist in the table. \\
e11 & Invalid:Alias & Used same column name in a single statement without any alias. \\
e12 & Group-by:Not Detected & Didn’t use GROUP BY keyword where it should be used. \\
e13 & Group-by:Wrong Cols & Group by wrong columns or unnecessary group by.        
\end{tblr}
}
\end{table}

In our error analysis (\cref{c:erroranalysis}), we manually identified various \textbf{Error Types} and outlined them in Table~\ref{tab:errtype}, each comprising multiple \textbf{correction cases}. These cases provide details like the error type, correct SQL, reasons for failure, and instructions. Additionally, we transform 'Case' information from these cases into embedding vectors using the Sentence Transformer model \cite{reimers2019sentence}, storing them in FAISS \cite{johnson2019billion} for similarity searches with new error case query vectors.

\begin{figure*}[!ht]  
\centerline{\includegraphics[width=\textwidth]{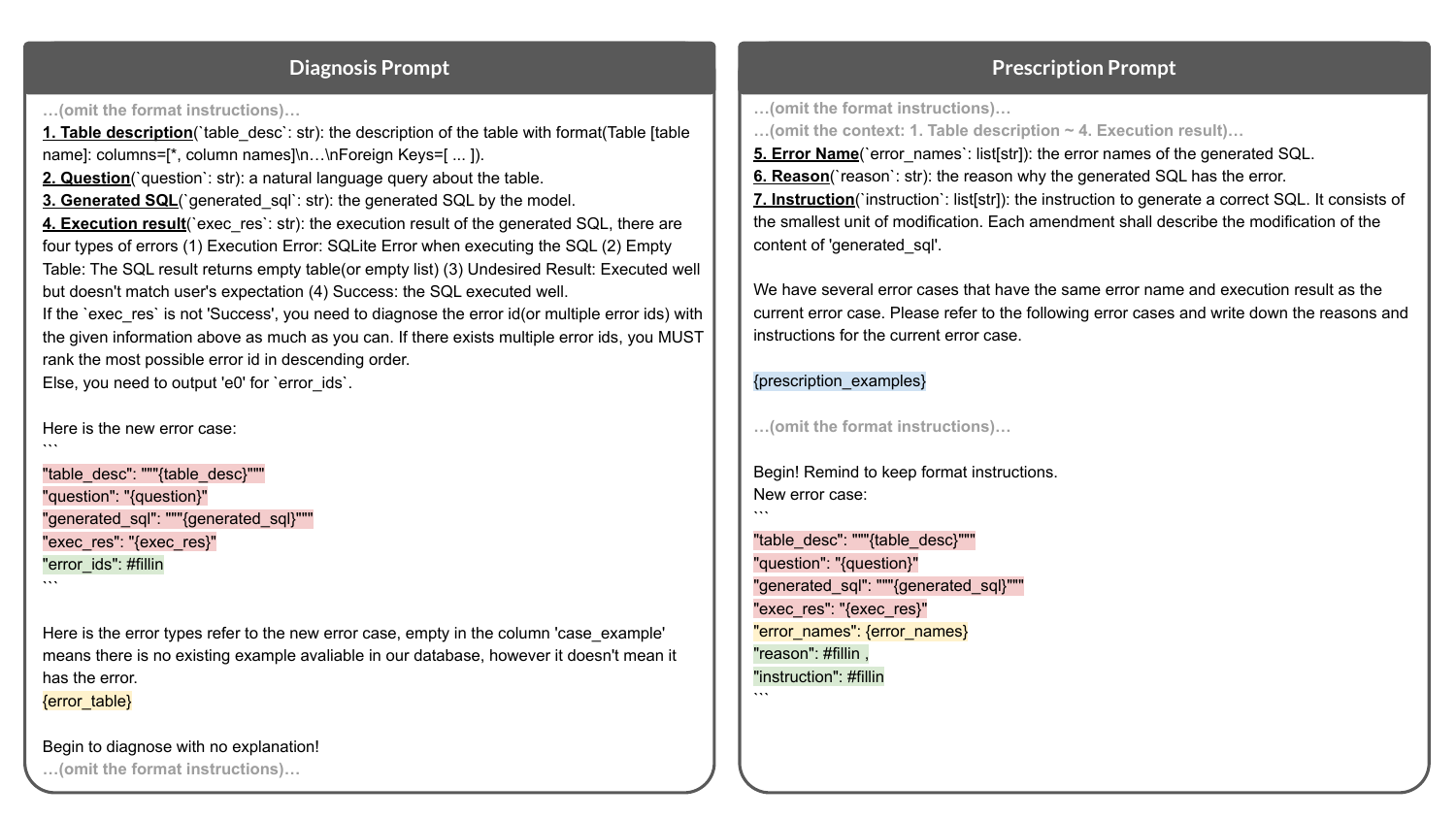}}
\caption{Diagnosis Prompt and Prescription Prompt are designed to diagnose and write prescriptions. Red blocks refer to the new error case information, yellow blocks are error type information, blue blocks are relevant case examples, and green blocks are answers that LLM should fill in.}
\label{fig:prompts}
\end{figure*}
% \vspace{-10pt}

\noindent\textbf{Step 1: Diagnosing.}
LLM performs well in the classification problem with a diagnostic reasoning process \cite{sun2023text}. We developed a \textbf{Diagnosis Prompt} for LLMs to classify error types. When an error arises, case details and an error types table are input as text to the prompt (indicated by red and yellow lines in Fig.~\ref{fig:prompts}). The LLM outputs one or more error types, ranked in descending order of severity, highlighting the most critical error first.

\noindent\textbf{Step 2: Writing Prescription.} 
In our experiment (\cref{c:experiments}), using the self-generic correction prompt from \cite{pourreza2023din}, we found that LLMs can't correct SQL errors due to limited knowledge. To address this, we introduced a \textbf{Prescription Prompt} for LLMs. This prompt generates reasoning and instructions for the new error case. It starts by converting the error case into structured text and using the Sentence Transformer model to embed it as a vector. Then, it finds the most relevant cases from the vector database via similarity searching and transforms them into text for the Prescription Prompt, as shown in the blue blocks of Fig.~\ref{fig:prompts}.

\noindent\textbf{Step 3: Applying Treatment.}
Given the instructions on fixing the generated SQL, the final step is to serve it as input to the \textbf{Treatment Prompt} for LLM to generate the correct SQL. The system will attempt this up to three times until a `Success' is achieved in the execution result.

\subsection{Embedding Fine-Tuning}\label{cc:eft}

Since LLMs are trained on vast and diverse data sources, they can understand diverse contexts. Intuitively, words with similar meanings are closer to each other in the embedding space, and those that are not are further apart. However, in our case, embedding vectors for correction cases should be organized by case types rather than token meanings. 
Thus, we fine-tuned a Sentence Transformer model with MPNet \cite{song2020mpnet} base architecture using a dataset of correction cases labeled with 14 labels (13 error types and one success). We employed triplet loss \cite{hoffer2015deep} for 20 epochs to enhance relevance in case retrieval.

% the embedding space in our scenario is a collection of cases. The semantic meaning should be organized based on the success or error types rather than the meaning of tokens composed in the context. Therefore, it is necessary to fine-tune an embedding model to better retrieve relevant cases in step 2. We built a dataset using correction cases with 13 error types as labels, then used triplet loss \cite{hoffer2015deep} to fine-tune a pre-trained Sentence Transformer model with MPNet \cite{song2020mpnet} as a base architecture for 20 epochs.

\section{Experiments and Results}\label{c:experiments}

\subsection{Evaluations, Models, and Metrics}

Our evaluation was conducted on the Spider \cite{yu2018spider} development set, where ground-truth queries can be easily accessible. We used OpenAI's models(GPT3.5-turbo, GPT4-turbo) with 0.01 temperature and a different number of max tokens(100, 1,024, and 600 for each step in ECPT). The performance was measured by execution accuracy in most of the experiments. To assess the effectiveness of the diagnosis, we utilize a straightforward hit rate metric: the number of trials that succeed in fixing errors divided by the number of total trials.

\subsection{Embedding Fine-Tuning}

\begin{figure*}[!hb]
\includegraphics[width=\textwidth]{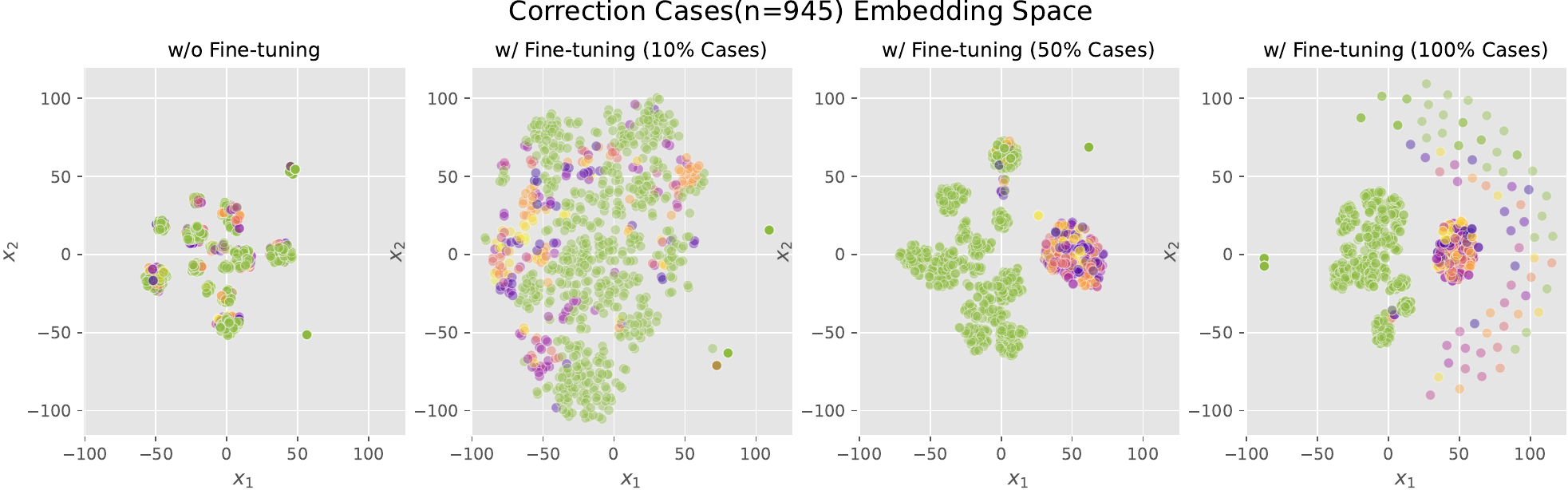}
\caption{Embedding vector spaces of correction cases visualized via T-SNE\cite{van2008visualizing}. Each point represents an embedded vector for each correction case. Success cases are colored green, and other colors are failed cases.}
\label{fig:embedding}
\end{figure*}

Optimizing tasks with RAG (Retriever-Augmented Generation) involves complex challenges, particularly in providing relevant examples and augmentations. We aim to enhance the semantic understanding of success and error types by refining the embedded vectors of correction cases, as illustrated in our results (Fig.~\ref{fig:embedding}). We evaluated the impact of fine-tuning on RAG by testing various sizes of correction cases, ensuring an equal distribution of each error type. This process revealed that while additional correction cases help distinctly separate success and error values in the embedding space, error cases remain closely clustered due to their multiple types.

\subsection{Results on Execution Accuracy}

\begin{figure*}[!h]
\centerline{\includegraphics[width=\textwidth]{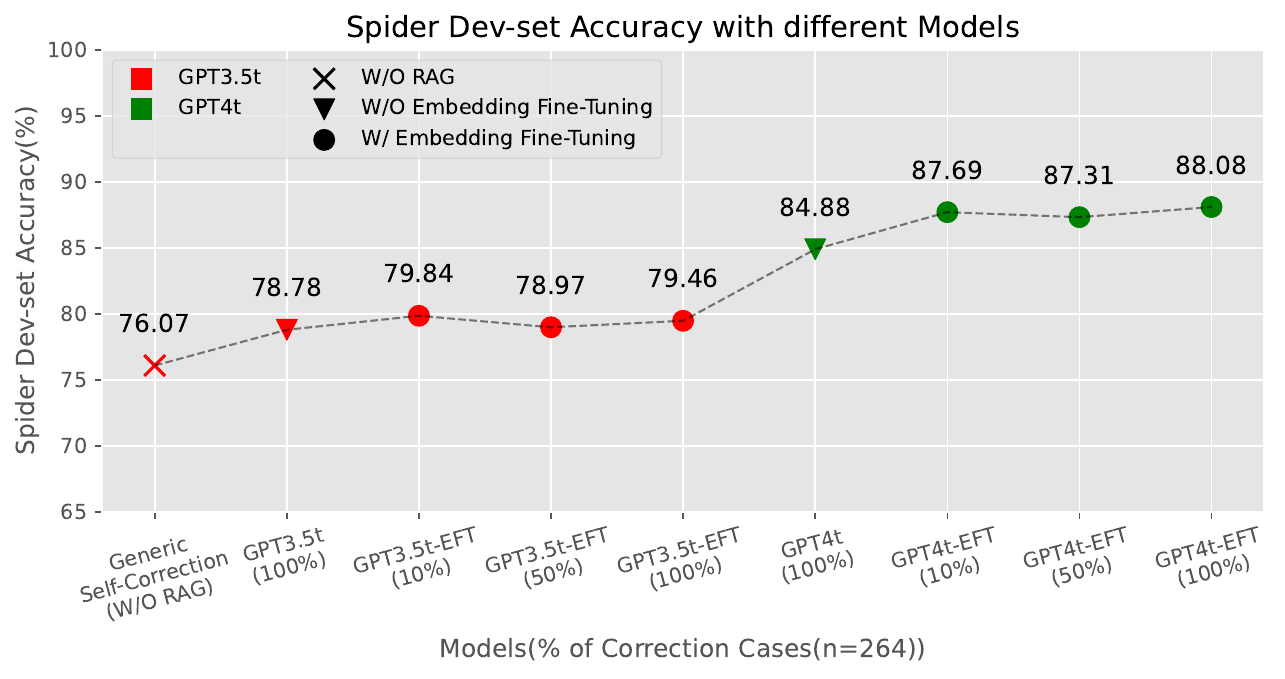}}
\caption{Experiments results. Percentage values in the x-axis mean accuracies of different configurations with LLM models. }
\label{fig:exp}
\end{figure*}

Fig.~\ref{fig:exp} shows that using different models and case numbers for RAG significantly improves execution accuracy on the development set. Switching to GPT4-turbo and incorporating an embedding fine-tuned model yielded about a 12\% gain compared to the baseline, the GPT3.5-turbo model with a generic self-correction prompt and no RAG. Both GPT3.5-turbo and GPT4-turbo models saw improvements with RAG from 76.07\% to 78.78\% and from 83.04\%(GPT4-turbo in Table \ref{tab:exp1}) to 84.88\%, respectively. Also, further benefits were gained from the embedding fine-tuned model (increasing by about 1\% and 3\%). Interestingly, the size of the correction cases had minimal impact, suggesting that correcting SQL errors might require fewer examples. This aspect needs more exploration in the future.

\begin{table}[!h]
\centering
\caption{Spider Dev-Set Performance Summary. Correction accuracy is determined by dividing successful fixes by 247 error cases. Execution accuracy is obtained by dividing the total of fixed cases and zero-shot prompting successes by 1,032 cases.}
\label{tab:exp1}
\resizebox{\linewidth}{!}{%
\begin{tblr}{
  cells = {c},
  row{1} = {t},
  hline{1,13} = {-}{0.08em},
  hline{2} = {-}{0.05em},
}
\textbf{Prompt Type} & \textbf{Model} & {\textbf{Option A}\\ \textbf{Fine-Tuned }\\ \textbf{Embeddings}} & {\textbf{Option B}\\ \textbf{Provide Example }\\\textbf{in Diagnosis}} & {\textbf{Option C}\\ \textbf{Resolve all }\\\textbf{at once}} & {\textbf{Correction }\\\textbf{Accuracy }\\\textbf{(247 cases)}} & {\textbf{Spider Dev-set }\\\textbf{Execution Accuracy }\\\textbf{(1032 cases)}} \\
- & GPT4 & - & - & -  & - & 77.33\% \\
- & GPT4-turbo & - & - & -  & - & 83.04\% \\
Generic & GPT3.5-turbo   & False & False & False   & 0.00\% & 76.07\%  \\
ECPT     & GPT3.5-turbo   & False & False & False   & 11.34\% & 78.78\%  \\
ECPT     & GPT3.5-turbo   & True & False & False   & 14.17\% & 79.46\%  \\
ECPT     & GPT3.5-turbo   & True  & True & False   & 14.57\% & 79.55\%  \\
ECPT     & GPT4-turbo     & False & False & False   & 36.84\% & 84.88\%  \\
ECPT     & GPT4-turbo     & True  & False & False   & 50.20\% & 88.08\%  \\
ECPT     & GPT4-turbo     & True  & True & False   & 44.13\% & 86.63\%  \\
ECPT     & GPT4-turbo     & True  & False & True    & 48.18\% & 87.60\%  \\
ECPT     & GPT4-turbo     & True  & True & True & 50.61\% & 88.18\%
\end{tblr}
}
\end{table}

Table~\ref{tab:exp1} presents the execution accuracy in an ablation study. Zero-shot NL-to-SQL results are shown in the first two rows. Option A uses a fine-tuned embedding model, while Option B employs a diagnostic prompt with a new error case alongside each error type. Option C selects only the top error types post-diagnosis. Embedding fine-tuning improved error correction, but other options didn't enhance performance. Compared with other works as a reference, Din-SQL \cite{pourreza2023din} achieved 74.2\% execution accuracy on the dev set of Spider. Our approach emphasizes error correction and needs feedback based on ground-truth SQL queries, especially for execution result: ``Undesired Result.'' 

\subsection{Hit Rate and Cost Usage}

Table \ref{tab:expcost} reports the hit rate, token usage, and costs while using API, a novel analysis not previously undertaken. We investigate the cost-effectiveness of accuracy improvements in three key experiments. The accuracy gain per dollar spent compared to the baseline (76.07\%) is 0.52\%, 0.59\%, and 0.38\% for each experiment. Notably, despite its higher hit rate, the third experiment incurs higher costs due to excessive token usage in input prompts, approximately 1.6 times more than the GPT4t-EFT experiment.

\begin{table}
\centering
\caption{Hit Rate and Cost Usage. Prompt tokens are the number of tokens used for input prompts. Completion tokens are the number of tokens that LLM generates.}
\label{tab:expcost}
\resizebox{\linewidth}{!}{%
\begin{tblr}{
  row{1} = {c},
  cell{2}{2} = {c},
  cell{2}{3} = {c},
  cell{2}{4} = {c},
  cell{2}{5} = {c},
  cell{2}{6} = {c},
  cell{2}{7} = {c},
  cell{3}{2} = {c},
  cell{3}{3} = {c},
  cell{3}{4} = {c},
  cell{3}{5} = {c},
  cell{3}{6} = {c},
  cell{3}{7} = {c},
  cell{4}{2} = {c},
  cell{4}{3} = {c},
  cell{4}{4} = {c},
  cell{4}{5} = {c},
  cell{4}{6} = {c},
  cell{4}{7} = {c},
  hline{1,5} = {-}{0.08em},
  hline{2} = {-}{0.05em},
}
Experiments & Prompt Tokens & Completion Tokens & Total Cost(\$) & Hit Rate & \# of Trials & Execution Accuracy\\
GPT3.5t-EFT & 2,020,555 & 129,455 & 6.58 & 5.09\% & 687 & 79.46\%\\
GPT4t-EFT & 1,665,553 & 119,112 & 20.23 & 23.01\% & 539 & 88.08\%\\
GPT4t-EFT w/ options B, C & 2,808,228 & 120,913 & 31.71 & 23.81\% & 525 & 88.18\%
\end{tblr}
}
\end{table}
\section{Conclusion}\label{c:conclusion}

This paper proposed a novel approach to NL-to-SQL, addressing a critical gap in current systems. By integrating error correction with prompt tuning, embedding fine-tuning, and RAG, we addressed the crucial need for accurate translation of natural language questions into SQL expressions. Our method drew inspiration from medical diagnosis processes and 
went beyond simple query translation; it intelligently diagnoses and corrects errors, leveraging external knowledge bases to refine its outputs. The notable 12\% accuracy improvement over existing baselines underscores the effectiveness of our framework, marking a substantial advancement in data access and management. This breakthrough has far-reaching implications, offering a powerful tool for diverse users, particularly in decision-making roles, and sets a new benchmark for future research and development in NL-to-SQL.

We see much room for improvement in our framework. First, since LLM-generated queries need to be verified with ground-truth ones, we can integrate Human-in-the-loop approaches to address this challenge. Second, manual error type initialization can be automatic by utilizing LLM agents. Finally, RAG-based prompt tuning remains resource-intensive due to the nature of the decomposed error correction process.
\newpage
\bibliographystyle{unsrt} % splncs04
\bibliography{bibliography}

@misc{androutsopoulos1995natural,
      title={Natural Language Interfaces to Databases - An Introduction}, 
      author={I. Androutsopoulos and G. D. Ritchie and P. Thanisch},
      year={1995},
      eprint={cmp-lg/9503016},
      archivePrefix={arXiv},
      primaryClass={cmp-lg}
}

@article{woods1972lunar,
  title={The lunar sciences natural language information system},
  author={Woods, William},
  journal={BBN report},
  year={1972},
  publisher={Bolt Beranek and Newman}
}

@article{saha2016athena,
  title={ATHENA: an ontology-driven system for natural language querying over relational data stores},
  author={Saha, Diptikalyan and Floratou, Avrilia and Sankaranarayanan, Karthik and Minhas, Umar Farooq and Mittal, Ashish R and {\"O}zcan, Fatma},
  journal={Proceedings of the VLDB Endowment},
  volume={9},
  number={12},
  pages={1209--1220},
  year={2016},
  publisher={VLDB Endowment}
}

@article{hwang2019comprehensive,
  title={A comprehensive exploration on wikisql with table-aware word contextualization},
  author={Hwang, Wonseok and Yim, Jinyeong and Park, Seunghyun and Seo, Minjoon},
  journal={arXiv preprint arXiv:1902.01069},
  year={2019}
}

@article{lyu2020hybrid,
  title={Hybrid ranking network for text-to-sql},
  author={Lyu, Qin and Chakrabarti, Kaushik and Hathi, Shobhit and Kundu, Souvik and Zhang, Jianwen and Chen, Zheng},
  journal={arXiv preprint arXiv:2008.04759},
  year={2020}
}

@article{lin2020bridging,
  title={Bridging textual and tabular data for cross-domain text-to-SQL semantic parsing},
  author={Lin, Xi Victoria and Socher, Richard and Xiong, Caiming},
  journal={arXiv preprint arXiv:2012.12627},
  year={2020}
}

@article{wang2019rat,
  title={Rat-sql: Relation-aware schema encoding and linking for text-to-sql parsers},
  author={Wang, Bailin and Shin, Richard and Liu, Xiaodong and Polozov, Oleksandr and Richardson, Matthew},
  journal={arXiv preprint arXiv:1911.04942},
  year={2019}
}

@article{radford2019language,
  title={Language models are unsupervised multitask learners},
  author={Radford, Alec and Wu, Jeffrey and Child, Rewon and Luan, David and Amodei, Dario and Sutskever, Ilya and others},
  journal={OpenAI blog},
  volume={1},
  number={8},
  pages={9},
  year={2019}
}

@article{brown2020language,
  title={Language models are few-shot learners},
  author={Brown, Tom and Mann, Benjamin and Ryder, Nick and Subbiah, Melanie and Kaplan, Jared D and Dhariwal, Prafulla and Neelakantan, Arvind and Shyam, Pranav and Sastry, Girish and Askell, Amanda and others},
  journal={Advances in neural information processing systems},
  volume={33},
  pages={1877--1901},
  year={2020}
}

@article{ouyang2022training,
  title={Training language models to follow instructions with human feedback},
  author={Ouyang, Long and Wu, Jeffrey and Jiang, Xu and Almeida, Diogo and Wainwright, Carroll and Mishkin, Pamela and Zhang, Chong and Agarwal, Sandhini and Slama, Katarina and Ray, Alex and others},
  journal={Advances in Neural Information Processing Systems},
  volume={35},
  pages={27730--27744},
  year={2022}
}

@article{rajkumar2022evaluating,
  title={Evaluating the text-to-sql capabilities of large language models},
  author={Rajkumar, Nitarshan and Li, Raymond and Bahdanau, Dzmitry},
  journal={arXiv preprint arXiv:2204.00498},
  year={2022}
}

@article{liu2023pre,
  title={Pre-train, Prompt, and Predict: A Systematic Survey of Prompting Methods in Natural Language Processing},
  author={Liu, Pengfei and Yuan, Weizhe and Fu, Jinlan and Jiang, Zhengbao and Hayashi, Hiroaki and Neubig, Graham},
  journal={arXiv preprint arXiv:2107.13586},
  year={2021}
}

@article{yu2018spider,
  title={Spider: A large-scale human-labeled dataset for complex and cross-domain semantic parsing and text-to-sql task},
  author={Yu, Tao and Zhang, Rui and Yang, Kai and Yasunaga, Michihiro and Wang, Dongxu and Li, Zifan and Ma, James and Li, Irene and Yao, Qingning and Roman, Shanelle and others},
  journal={arXiv preprint arXiv:1809.08887},
  year={2018}
}

@article{pourreza2023din,
  title={Din-sql: Decomposed in-context learning of text-to-sql with self-correction},
  author={Pourreza, Mohammadreza and Rafiei, Davood},
  journal={arXiv preprint arXiv:2304.11015},
  year={2023}
}

@article{chowdhery2022palm,
  title={Palm: Scaling language modeling with pathways},
  author={Chowdhery, Aakanksha and Narang, Sharan and Devlin, Jacob and Bosma, Maarten and Mishra, Gaurav and Roberts, Adam and Barham, Paul and Chung, Hyung Won and Sutton, Charles and Gehrmann, Sebastian and others},
  journal={arXiv preprint arXiv:2204.02311},
  year={2022}
}

@article{touvron2023llama,
  title={Llama: Open and efficient foundation language models},
  author={Touvron, Hugo and Lavril, Thibaut and Izacard, Gautier and Martinet, Xavier and Lachaux, Marie-Anne and Lacroix, Timoth{\'e}e and Rozi{\`e}re, Baptiste and Goyal, Naman and Hambro, Eric and Azhar, Faisal and others},
  journal={arXiv preprint arXiv:2302.13971},
  year={2023}
}

@article{wu2023bloomberggpt,
  title={Bloomberggpt: A large language model for finance},
  author={Wu, Shijie and Irsoy, Ozan and Lu, Steven and Dabravolski, Vadim and Dredze, Mark and Gehrmann, Sebastian and Kambadur, Prabhanjan and Rosenberg, David and Mann, Gideon},
  journal={arXiv preprint arXiv:2303.17564},
  year={2023}
}

@article{sun2023text,
  title={Text Classification via Large Language Models},
  author={Sun, Xiaofei and Li, Xiaoya and Li, Jiwei and Wu, Fei and Guo, Shangwei and Zhang, Tianwei and Wang, Guoyin},
  journal={arXiv preprint arXiv:2305.08377},
  year={2023}
}

@article{lewis2021retrievalaugmented,
  title={Retrieval-Augmented Generation for Knowledge-Intensive NLP Tasks},
  author={Lewis, Patrick and Perez, Ethan and Piktus, Aleksandra and Petroni, Fabio and Karpukhin, Vladimir and Goyal, Naman and K{\"u}ttler, Heinrich and Lewis, Mike and Yih, Wen-tau and Rockt{\"a}schel, Tim and others},
  journal={arXiv preprint arXiv:2005.11401},
  year={2020}
}

@article{wei2022chain,
  title={Chain-of-Thought Prompting Elicits Reasoning in Large Language Models},
  author={Wei, Jason and Wang, Xuezhi and Schuurmans, Dale and Bosma, Maarten and Ichter, Brian and Xia, Fei and Chi, Ed and Le, Quoc and Zhou, Denny},
  journal={arXiv preprint arXiv:2201.11903},
  year={2022}
}

@article{gao2023text,
  title={Text-to-SQL Empowered by Large Language Models: A Benchmark Evaluation},
  author={Gao, Dawei and Wang, Haibin and Li, Yaliang and Sun, Xiuyu and Qian, Yichen and Ding, Bolin and Zhou, Jingren},
  journal={arXiv preprint arXiv:2308.15363},
  year={2023}
}

@article{van2008visualizing,
  title={Visualizing data using t-SNE.},
  author={Van der Maaten, Laurens and Hinton, Geoffrey},
  journal={Journal of machine learning research},
  volume={9},
  number={11},
  year={2008}
}

@article{reimers2019sentence,
  title={Sentence-bert: Sentence embeddings using siamese bert-networks},
  author={Reimers, Nils and Gurevych, Iryna},
  journal={arXiv preprint arXiv:1908.10084},
  year={2019}
}

@article{johnson2019billion,
  title={Billion-scale similarity search with gpus},
  author={Johnson, Jeff and Douze, Matthijs and J{\'e}gou, Herv{\'e}},
  journal={IEEE Transactions on Big Data},
  volume={7},
  number={3},
  pages={535--547},
  year={2019},
  publisher={IEEE}
}

@inproceedings{li2022resdsql,
  author = {Haoyang Li and Jing Zhang and Cuiping Li and Hong Chen},
  title = "RESDSQL: Decoupling Schema Linking and Skeleton Parsing for Text-to-SQL",
  booktitle = "AAAI",
  year = "2023"
}

@article{scholak2021picard,
  title={PICARD: Parsing incrementally for constrained auto-regressive decoding from language models},
  author={Scholak, Torsten and Schucher, Nathan and Bahdanau, Dzmitry},
  journal={arXiv preprint arXiv:2109.05093},
  year={2021}
}

@misc{song2205comprehensive,
  title={A comprehensive survey of few-shot learning: Evolution, applications, challenges, and opportunities (2022)},
  author={Song, Yisheng and Wang, Ting and Mondal, SK and Sahoo, JP},
  journal={arXiv preprint arxiv:2205.06743}
}

@article{hoffer2015deep,
  title={Deep metric learning using Triplet network},
  author={Hoffer, Elad and Ailon, Nir},
  journal={arXiv preprint arXiv:1412.6622},
  year={2014}
}

@article{song2020mpnet,
  title={Mpnet: Masked and permuted pre-training for language understanding},
  author={Song, Kaitao and Tan, Xu and Qin, Tao and Lu, Jianfeng and Liu, Tie-Yan},
  journal={Advances in Neural Information Processing Systems},
  volume={33},
  pages={16857--16867},
  year={2020}
}




% LNCS Homepage, \url{http://www.springer.com/lncs}. Last accessed 4
% \end{thebibliography}
\end{document}